**Beyond the Proving Ground: Independent Public-Road Testing of Assisted Lane Change Systems using LiDAR**

**Marcello Cellina***
European Commission Joint Research Centre
Email: name.surname@ec.europa.eu

**Akos Kriston**
European Commission Joint Research Centre
Email: name.surname@ec.europa.eu

**Antonio Migneco**
European Commission Joint Research Centre
Email: name.surname@ec.europa.eu

**Davide Maggi**
European Commission Joint Research Centre
Email: name.surname@ec.europa.eu

**Stefano Favelli**
European Commission Joint Research Centre
Email: name.surname@ec.europa.eu

**Fabrizio Re**
European Commission Joint Research Centre
Email: name.surname@ec.europa.eu

**Fabrizio Minarini**
European Commission Joint Research Centre
Email: name.surname@ec.europa.eu

**Andrea Nuovo**
European Commission Joint Research Centre
Email: name.surname@ec.europa.eu

**Riccardo Dona**
European Commission Joint Research Centre
Email: name.surname@ec.europa.eu

**Biagio Ciuffo**
European Commission Joint Research Centre
Email: name.surname@ec.europa.eu

*Corresponding Author

## ABSTRACT

**Objectives:** Testing of commercial Advanced Driver Assistance Systems (ADAS) is essential to ensure safety and compliance during type approval and in-service operations. However, these systems are typically tested and certified on proving grounds, using scenarios that may not adequately reflect the complexity of real-world driving conditions. In addition, the geo-fencing of many systems often requires manufacturer collaboration for proving ground testing, further limiting the independence of the safety assessment. To address this limitation, this work presents a novel methodology for independently testing ADAS systems on public roads, with a specific focus on Assisted Lane Change (ALC) functionalities.

**Methods**: We designed and conducted a road-testing campaign in the A31 French motorway using an ALC-capable test vehicle equipped with a LiDAR-based Vehicle Detection and Tracking (VDT) system. We run a test matrix on public roads, covering multiple combinations of inter-vehicle distance and speeds between the Vehicle Under Test (VUT) and the Take-Over (TO) vehicle. Both vehicles were also equipped with RTK GNSS receivers, which were used to quantify the performance of the VDT system. The trajectories recorded during the ALC manoeuvres were then compared with the theoretical behaviour from the UNECE Regulation No. 79 (R79) to verify the execution of the suppression of the lane change procedure in real-word driving.

**Findings**: We initiated 27 predefined ALC manoeuvres in public road. Of these, 18 were completed successfully, while 9 were suppressed by the system. By comparing the results with the requirements prescribed by R79, we found that in 6 cases the system was allowing manoeuvres that does not respect the regulatory minimum distance requirements. Furthermore, in 3 of these cases, the deviation rema ined statistically significant after accounting for the measurement uncertainty of the instrumentation.

**Novelty**: To the authors' knowledge, this paper reports the results of the first public-road testing campaign specifically designed to evaluate the regulatory compliance of ALC systems with respect to UNECE Regulation No. 79 critical safety distance requirements. It also proves the effectiveness of a LiDAR-based sensor system and shows its suitability for this purpose.

**Practical Applications:** The methodology can support market surveillance of commercial vehicles type-approved under UNECE Regulation No. 79. The findings may also inform future revisions of R79 by highlighting real-world operating regions and behaviours not adequately addressed by existing approval procedures.

# INTRODUCTION

Vehicle automation has the potential to improve road safety and transport efficiency (Masello et al. 2022; Aleksa et al. 2024). To ensure that automated driving functionalities provide consistent levels of performance and safety, harmonized technical requirements and effective assessment procedures are essential. Within this context, UNECE regulations, together with the activities of Type Approval Authorities and Technical Services, play a central role by establishing common requirements and enabling the approval of vehicle systems across different markets. However, the effectiveness of this framework relies on the capabilities of the prescribed tests to adequately capture the operational behaviour and safety limitations of the evaluated system.

Current type-approval procedures are normally conducted under controlled conditions on proving grounds (Pietruch et al. 2020). Although this approach ensures repeatability and reproducibility, it necessarily covers limited scenarios which may not adequately represent the complexity of real-world driving. Consequently, narrow testing conditions have limited capability to ensure equivalent behaviour throughout the system's actual operating domain on public roads.

Furthermore, functions as Assisted Lane Change (ALC) are frequently geo-fenced to selected highways, requiring the manufacturer collaboration to be tested on proving grounds. While this is not a limitation for Type Approval, it becomes a strong barrier to independent vehicle testing for Market Surveillance activities. Additionally, the limited availability of Real-Time Kinematic Global Navigation Satellite System (RTK GNSS) signal complicates independent testing outside proving grounds with sufficent accuracy, particularly in tunnels, canyons, and rural areas with limited satellite visibility.

To address these limitations, this work presents the design and the results of a public-road testing campaign involving a commercial production vehicle equipped with an ALC system, using LiDAR as the main measurement instrument. A full factorial test matrix was implemented and executed by coordinating the Vehicle Under Test (VUT) and the approaching Take-Over (TO) vehicle on French motorways. To assess the behaviour of the ALC system, we instrumented the vehicles with LiDAR, cameras, GNSS receivers and dedicated storage equipment. The VDT offline processing of the LiDAR measurements allowed us to measure inter-vehicle distances and relative velocities during the ALC manoeuvres, and to compare the observed trajectories with the behaviour prescribed by R79 for the Suppression of the Lane Change Procedure. The RTK GNSS information was instead used only as a ground truth reference to characterize the LiDAR precision and detection rate and to evaluate the statistical significance of the results.

The main contribution of this paper is a novel methodology for independently testing ALC systems on public roads using LiDAR measurements instead of conventional proving-ground procedures and RTK GNSS-based instrumentation. The presented results pave the way for type approval and market surveillance under realistic driving conditions more easily, efficiently, and economically than conventional testing approaches.

## Related Works

Research on Assisted Lane-Change (ALC) systems includes both development-oriented validation and field experiments. (Ulbrich et al. 2017) present a scenario-based approach for testing tactical lane-change behaviour planning, whereas (Raboy et al. 2021) demonstrate a cooperative lane-change algorithm using Vehicle-to-Vehicle communication, vehicle-based radar, and automated longitudinal control in a proof-of-concept field experiment. More recently (Mattas et al. 2025) evaluated five commercial ALC systems in live traffic using GNSS/IMU and road camera data. Their safety analysis covered three vehicles and compared observed margins with UN Regulation No. 171, although it missed the study of system-suppressed manoeuvres.

Other studies focus on users, driver behaviour, or lane-change models rather than on regulatory compliance testing of production systems. For instance, (Noonan et al. 2022) analysed driver-initiated Tesla Auto Lane Changes from naturalistic in-vehicle data, but without any measurement of the interaction with the surrounding traffic. (Madigan et al. 2018) examined manual and automated overtaking in a driving simulator, while (Schakel et al. 2012) developed and calibrated a microscopic model of human lane-changing and car-following behaviour. Again, (Donà et al. 2026) extended a simulation-based safety reference model to include evasive lane-change responses in UN Regulation No. 157 scenarios; it did not experimentally test a production ALC function.

Broader reviews describe simulation, scenario-based methods, and on-road ADAS testing but do not provide a dedicated UN Regulation No. 79 ALC test procedure, as in (Khan et al. 2023; Tang et al. 2023). In addition, most of on-road ADAS studies have predominantly addressed longitudinal control: (Viti et al. 2008) investigated driver–ACC interaction in a field operational test, (Milanés et al. 2014) analysed Cooperative ACC controllers retrofitted on four production vehicles, while (Gunter et al. 2021) evaluated the string stability of seven commercial ACC systems.

Usage of LiDAR sensors, although widespread for Automated Driving Systems, recently saw an increase in its adoption for traffic safety assessment. For instance, works from (Bhattarai et al. 2024; Gargoum et al. 2022; Wu et al. 2020) use roadside LiDARs to detect and track vehicles and Vulnerable Road Users (VRUs), but they share the scalability limitations of every infrastructure-based system. Authors in (Kilani et al. 2021; Guerrero-Sevilla et al. 2025; Kim et al. 2025) use mobile LiDAR sensors, but their goal is the broader safety assessment, not type approval or market surveillance activities.

Therefore, despite growing interest in on-road ADAS testing and LiDAR-based assessment, the literature still lacks a public-road campaign explicitly evaluating ALC behaviour against UN Regulation No. 79 using vehicle-mounted LiDAR sensors. Moreover, existing work only contemplate completed ALC manoeuvres, while the study of system-suppressed attempts remains unaddressed.

## METHODS

### Critical Distance for Assisted Lane Change Manoeuvres

The UNECE Regulation No. 79 (UNECE 2023) describes a set of measurable requirements for each manoeuvre to ensure safe operations. In this work, we will focus on the Critical Distance between the VUT and TO, which must be respected to allow an ALC manoeuvre to be started, as described by the Suppression of the Lane Change Procedure requirements.

The Critical Distance is defined as a function of the speeds of the ALC vehicle and the approaching vehicle, as from

$$S_{crit} = \left(v_{app} - v_{ALC}\right) * t_B + \frac{\left(v_{app} - v_{ALC}\right)^2}{2 * a} + v_{ALC} * t_G, \qquad (1)$$

where $v_{app}$ and $v_{ALC}$ are, respectively, the velocities of the approaching and ALC vehicle, $a = 3m/s^2$ is the maximum approaching vehicle deceleration, $t_B = 0.4s$ is the reaction time of the approaching vehicle before its deceleration starts, and $t_G = 1s$ represents the minimum resulting time gap between the two vehicles after the manoeuvre has ended.

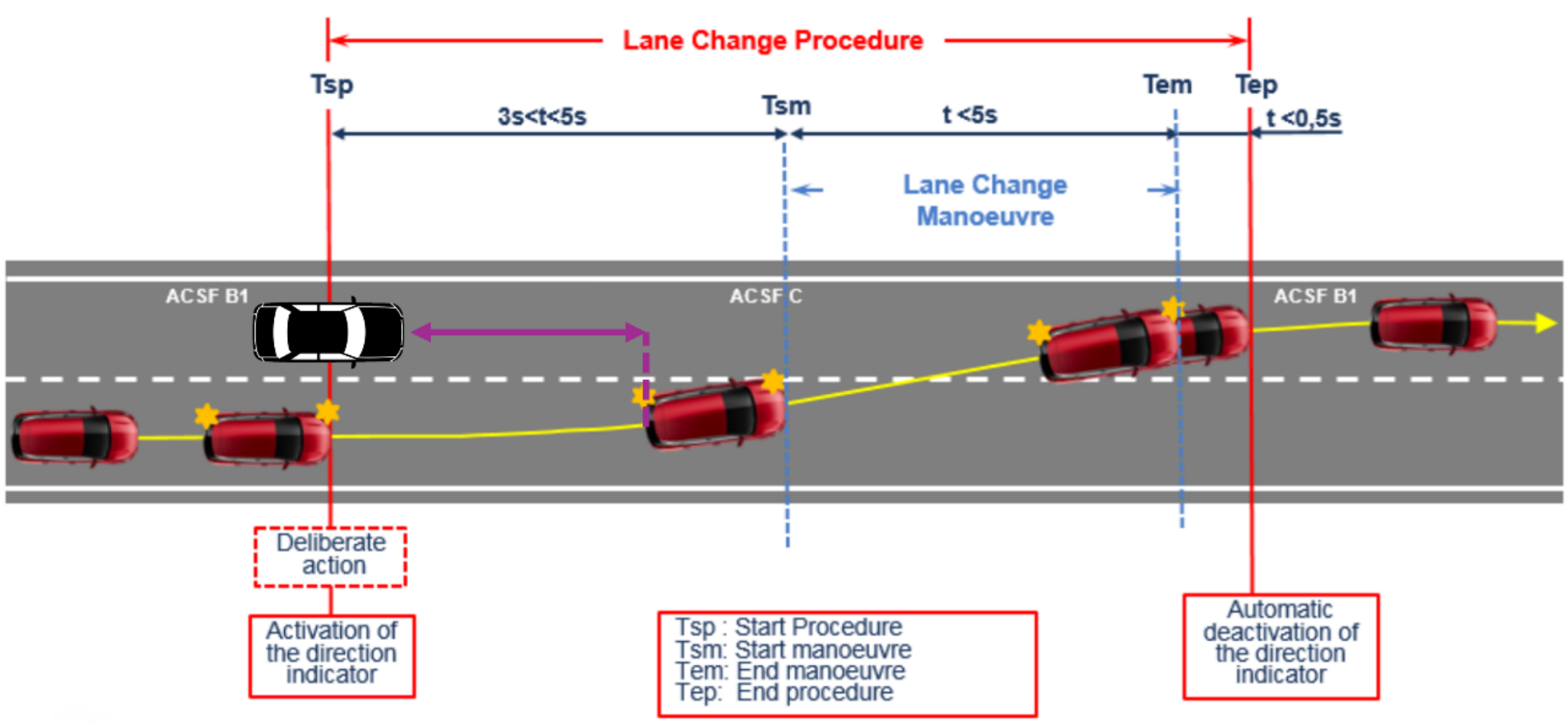

**Figure 1: Assisted Lane Change procedure as prescribed by R79. The Critical Distance (purple) between the VUT (red) and TO (black) vehicles must be respected at the Time of Start Measure ($T_{sm}$), when the VUT crosses the lane marking. Image adapted from (UNECE 2023)**

As per (UNECE 2023), the Critical Distance must be evaluated at the time of the lane marking crossing from the VUT during the Lane Change Procedure, as **Figure 1** shows. If the TO vehicle, approaching in the adjacent lane, is closer than the Critical Distance at the moment of the lane marking crossing from the VUT, the ALC system should suppress the manoeuvre.

In this work, we investigate if commercial UNECE Regulation No. 79 (R79) type approved ALC systems respect this limitation on public roads, meaning if they successfully reject ALC manoeuvres initiated with an approaching vehicle at a distance lower than the Critical Distance from R79.

**Experimental Setup and Testing Procedure**

The experimental campaign took place in March 2026, when we drove three type-approved passenger cars through the A31 French motorway, going North between Dijon and Nancy, as part of a broader public road-testing campaign.

The Vehicle Under Test (VUT), was equipped with R79 ACSF of category C Assisted Lane Change capabilities, which were tested before the start of the campaign and confirmed to be working on geo-fenced highway road stretches. The Take-Over (TO) vehicle, taking the role of the approaching vehicle from the regulation, was equipped with a rooftop-mounted LiDAR and camera system developed by IVEX N.V.  (Heverlee, Belgium) and was used as the approaching vehicle in the ALC manoeuvre as per R79 definitions, by staying behind VUT in the adjacent lane and following a target velocity, greater than VUT.

Additionally, a third vehicle, from now on referred to as “support”, did not participate directly in the experiment, but had the role of staying behind the VUT in the same lane to prevent any vehicle from inadvertently joining and invalidating the experiment. At the same time, it worked as reference position to trigger the ALC manoeuvre start at the desired distance by using its Advanced Cruise Control (ACC) system to maintain a fixed distance headway with respect to the VUT. **Figure 2** provides a graphical representation of this testing procedure.

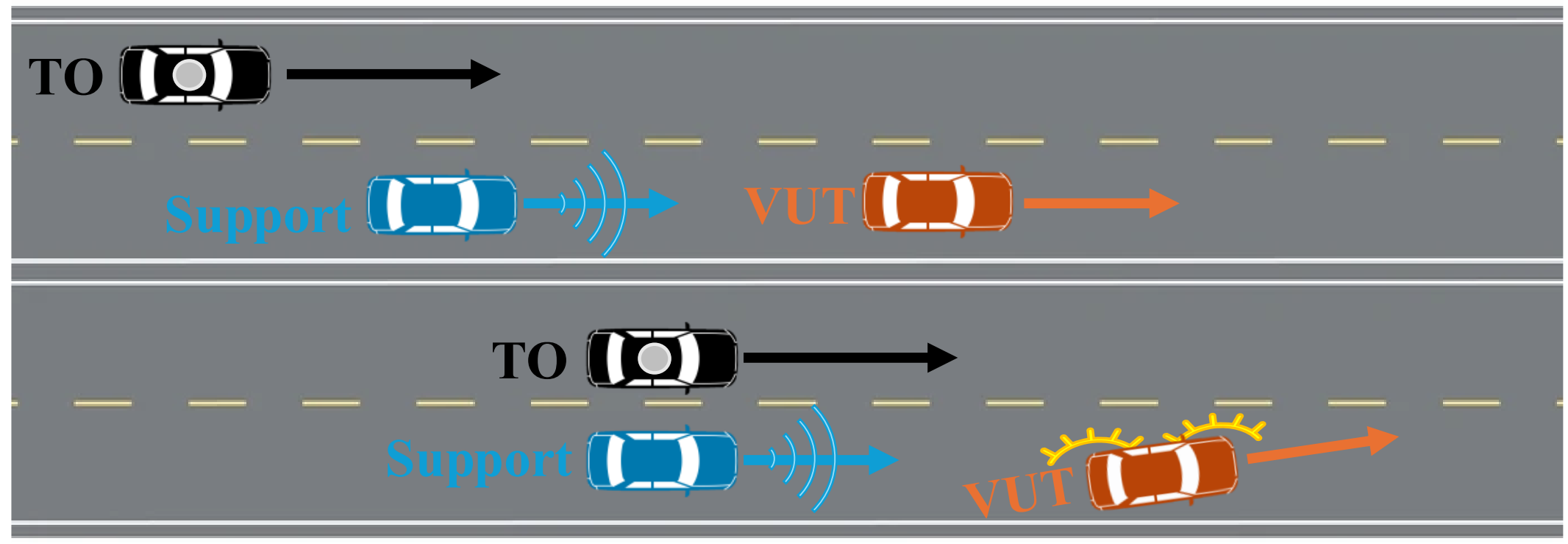

**Figure 2: The testing procedure: (top) the TO vehicle, equipped with the LiDAR (gray), is approaching the VUT (orange) and the support (blue). The support is using its ACC to keep a fixed distance from the VUT. (bottom) the Time Start Procedure (TSP): as soon as the TO is aligned with the support, the VUT activates the direction indicator and starts moving towards the lane marking.**

The VUT was equipped with a Septentrio AsteRx EB Dual-Antenna GNSS receiver, which was part of the commercial LiDAR VDT system, together with an Ouster OS1 128-layer LiDAR, while both the TO and the support vehicle mounted an OxTS RT1003 GNSS/INS receiver on board, for Ground Truth position computation. The commercial LiDAR system provides ego/vehicle and tracked object poses at 10 Hz, with position, velocity, heading, width and length estimation for each object.

The TO vehicle was also fitted with a dashboard camera, to determine the moment of first activation of the turn indicator, as well as an outside camera for lane detection, to determine the moment of lane crossing. Both these moments were annotated manually for each test, with a temporal resolution of one second.

With this setup, we run a full testing matrix where, for each combination of VUT and TO speeds, we triggered the ALC procedure start at three different distance values. These distances were selected to cover conditions close to both the R79 critical distance, as determined by the speeds of the VUT and TO, and the minimum activation distance specified in R79. This approach was specifically intended to assess whether the ALC system correctly discriminated between conditions in which the manoeuvre should be performed from those in which it should be suppressed. We repeated some combinations multiple times, especially those corresponding to critical situations as from R79. The combinations of speeds we could test were limited by the applicable speed limit, which on this highway stretch was 130 km/h.

## RESULTS

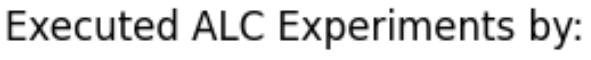

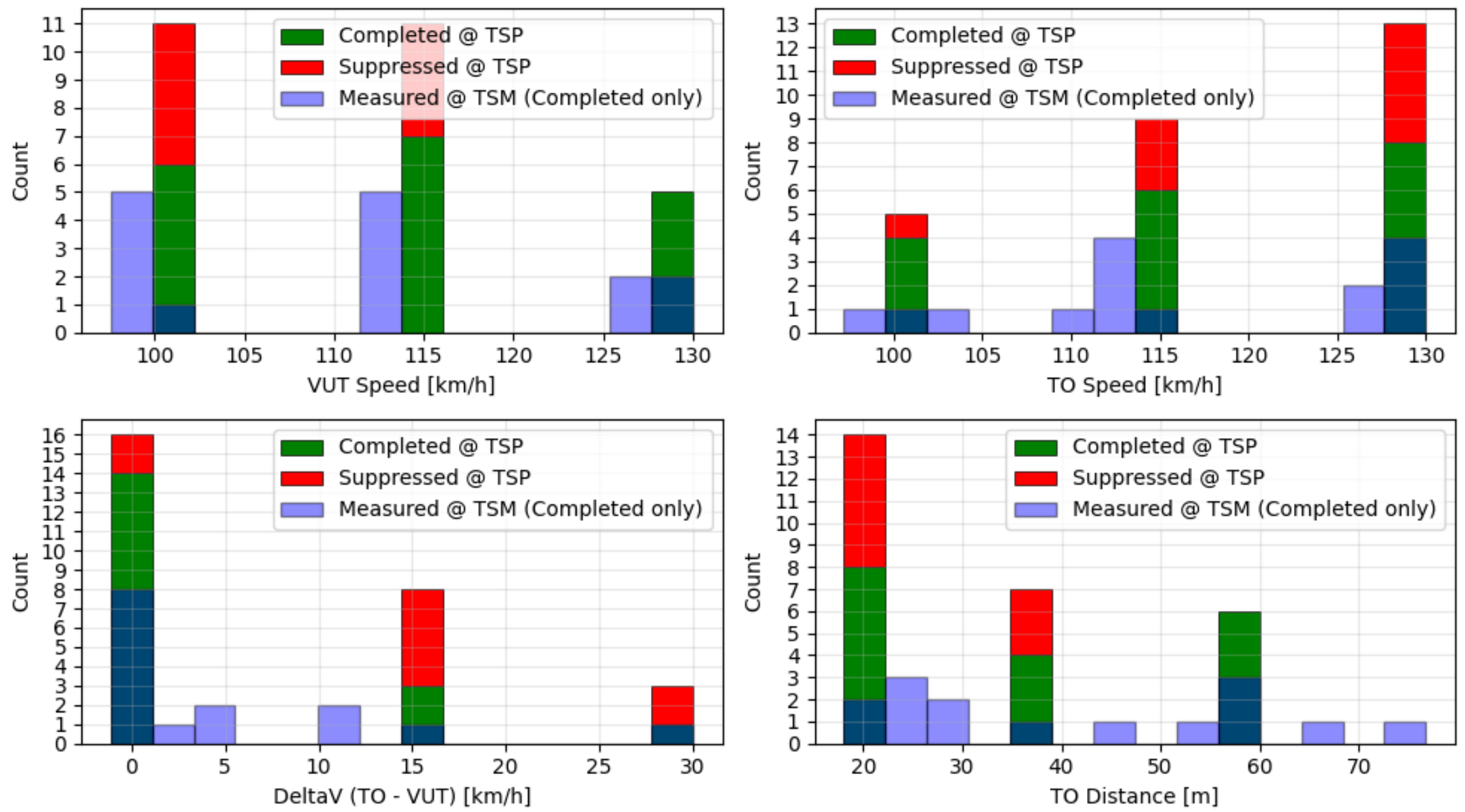

**Figure 3: Histogram of all ALC experiments and test matrix combinations. TSP =Time Start Procedure (turn signal activation), TSM = Time Start Manoeuvre (VUT Lane Crossing). The measured quantities at TSM are defined only for completed manoeuvres as suppression takes place before TSM.**

We recorded a total 27 ALC, of which 18 were successfully completed while the remaining 9 were suppressed. The tests encompassed speeds from 100 to 130km/h and triggering distances from 20 to 60m.

**Figure** 3 shows the distributions across the key indicators. While we were successful in tracking the desired VUT speed using ACC, we encountered technical issues in tracking the TO speed and the TLC distance, due to the non-constant delay between TSP and TSM.

We consistently triggered the ALC manoeuvre using the support vehicle as a distance marker, but we could not quantify the (non-constant) activation delay between the Time of Start Procedure (TSP) and the Time of Lane Crossing (TLC). At the same time, the ACC of the TO was already reacting to the lateral motion of the VUT, effectively reducing its speed.

While this procedure proved to be more accurate and repeatable compared to manual driving, procedural and technical aspects showed ample margin for improvement. Due to these shortcomings and to the limited detection range of the LiDAR, not all experiments in

**Figure** 3 have a value for the measured quantities at the TSM: of the Completed experiments, 3 were not tracked by the LiDAR, as they took place outside the LiDAR measurement range.

If we suppose for each experiment a constant VUT speed, which is a reasonable assumption considering that during the experiment the ACC was engaged, isolate each of the three values of VUT speed in the test matrix, for better understanding of the single lane change behaviour, as **Figure 4** shows. From this analysis, we can detect whether the R79 ALC suppression threshold is respected, as all the experiments in which at the Time of Lane Change (TLC), the TO vehicle is closer to VUT than the Critical Distance, but still the ALC manoeuvre is executed.

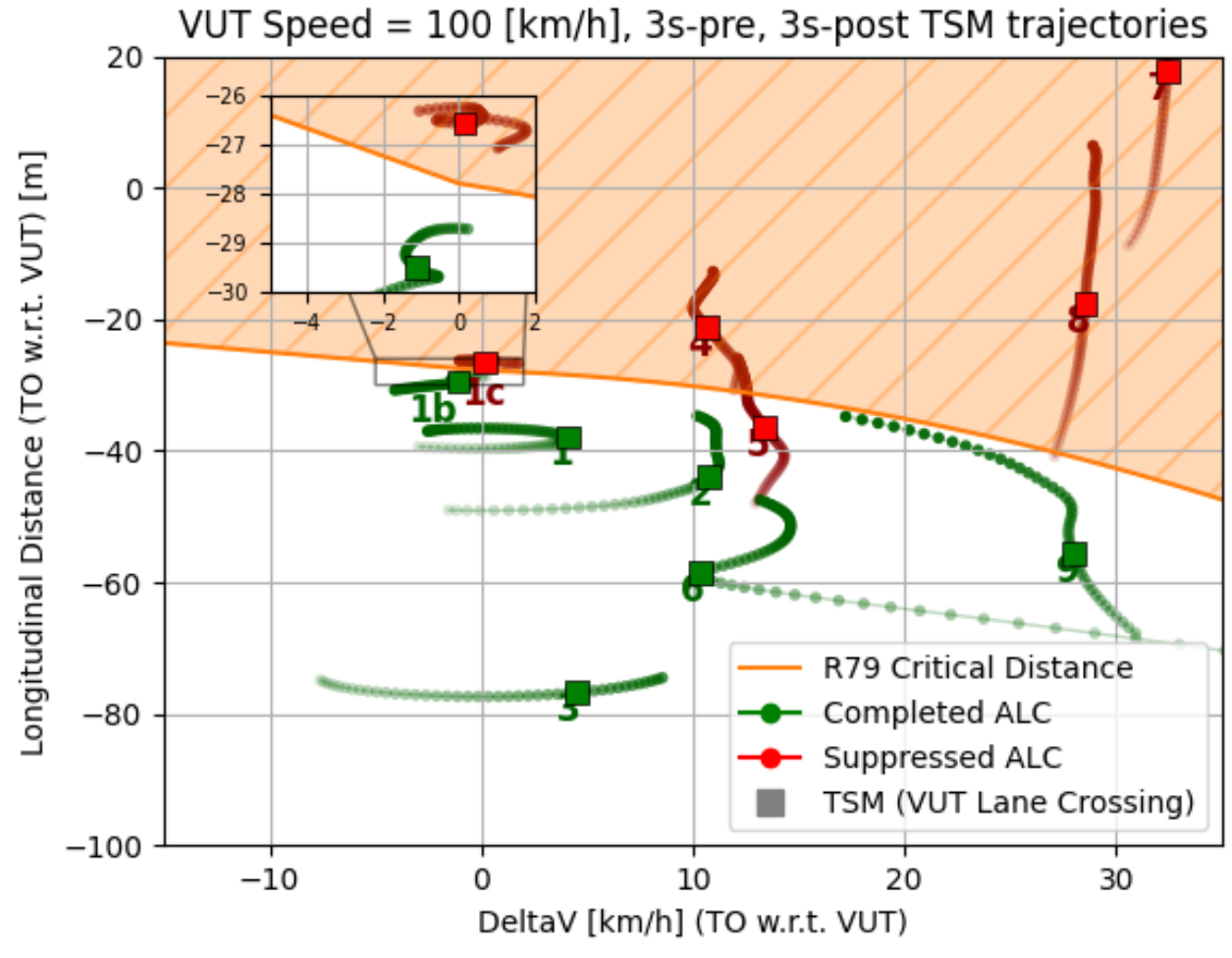

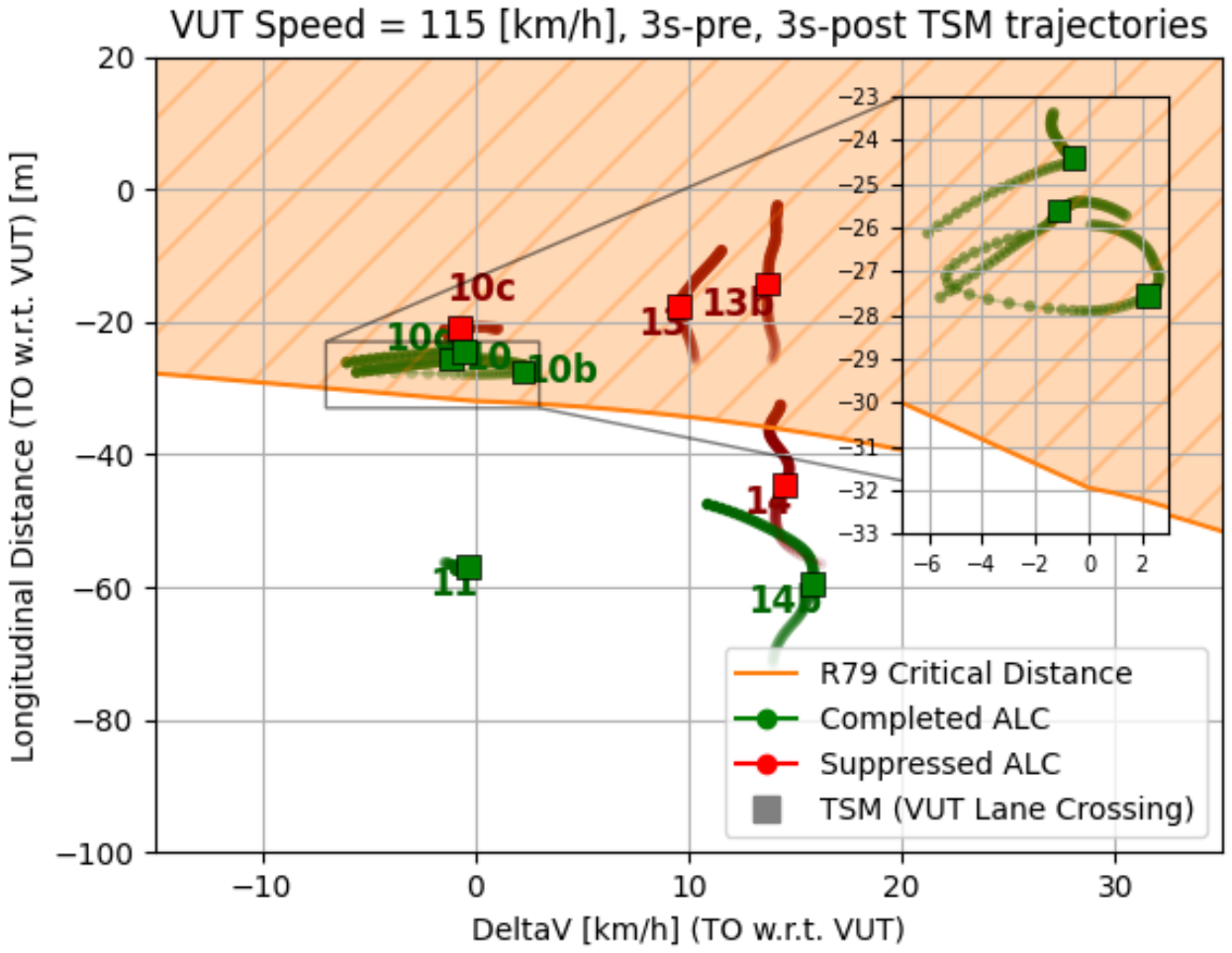

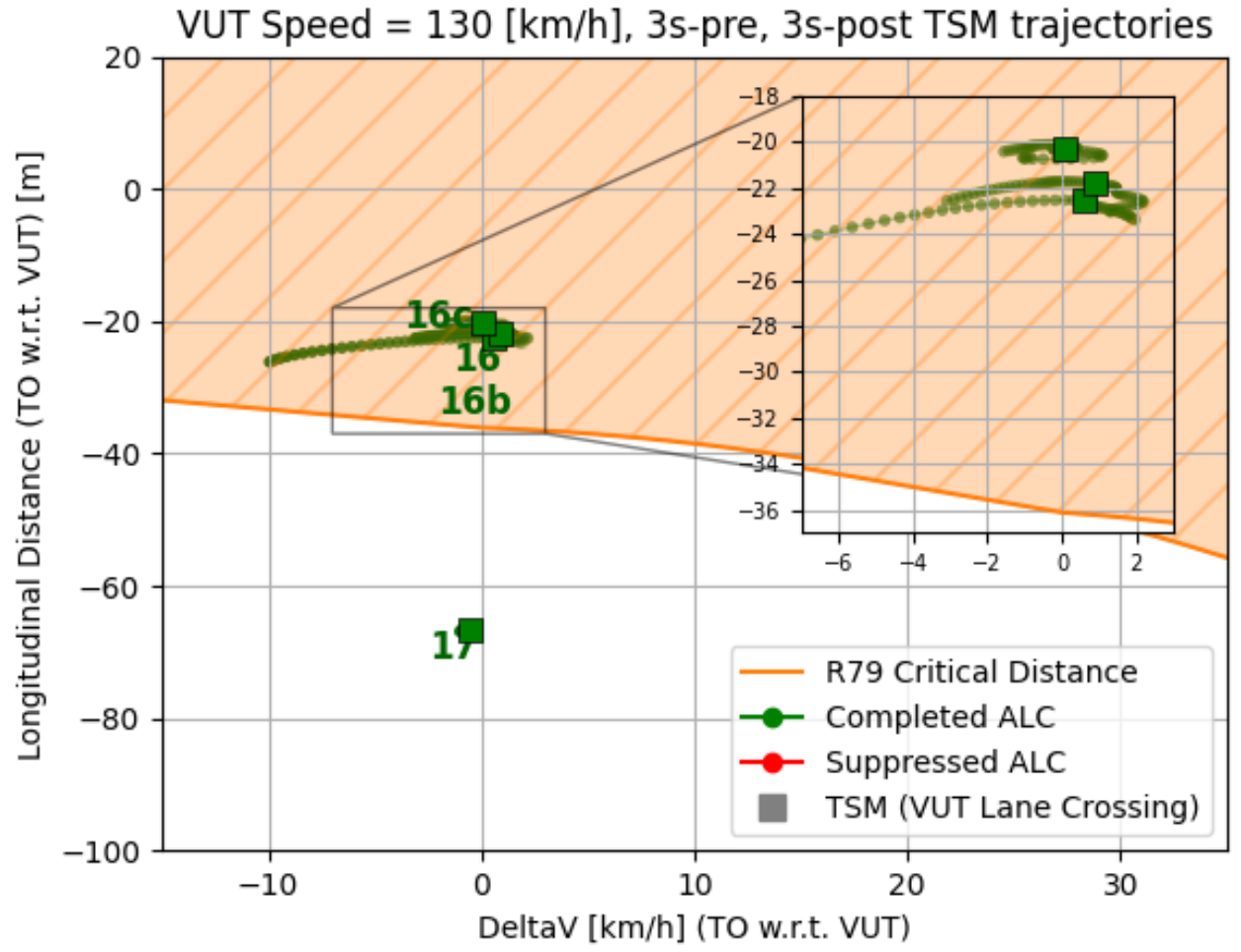

**Figure 4: Constant-VUT speed curves of the TO ALC trajectories, superimposed on the R79 critical distance. The distance is intended as bumper to bumper. The square dot represents TSM, the transparent markers represent the trajectory before the ALC.**

From **Figure 4,** we can distinguish the following conditions:

- **Critical Suppressed ALCs**, where at TLC the critical distance threshold is not respected and the manoeuvre is suppressed by the system. This is the intended, normal behaviour of the ALC. Examples are experiments 4,7,8,13 and 13b
- **Non-Critical Suppressed ALCs**, where the ALC system aborted or rejected a manoeuvre that would otherwise respect the Critical Distance condition, potentially due to other circumstances like driver attention monitoring or excessive precaution from the system, like experiments 7 and 14
- **Non-Critical Completed ALCs**, nominal behaviour for a completed ALC manoeuvre where the Critical Distance threshold is respected at TLC, like experiments 1,1b,2,3,6,9,11,14b and 17
- **Critical Completed ALCs: potential Critical Distance Overshoot** experiments, where the VUT performed the full ALC manoeuvre even if at TLC the distance with the TO vehicle was lower than the R79 Critical Distance. This happened in experiments 10, 10b, 10d, 16, 16b, 16c

According to the definitions form this paragraph, 33% of the completed ALC manoeuvres were initiated at TLC while the vehicles were at a shorter distance than the R79 Critical Distance. In the next section, we will evaluate if these results have statistical significance.

## DISCUSSION

In this section, we analyse the 6 experiments characterized by R79 Critical Distance Overshooting, as detected by the LiDAR, to determine the effect of the measurement noise on the assessment of the critical distance.

We used the RTK GNSS as the ground truth to characterize the LiDAR precision. However, it must be noted that RTK Integer Fix was only available as a solution type for both vehicles simultaneously only in the 2.3% of the recorded experimental data. LiDAR measurements have a much greater availability than RTK GNSS, although with less precision due to the hypotheses and approximations in the VDT pipeline.

The commercial LiDAR system used for this analysis was found to have a precision of the VDT position output quantified in 0.83 m (std) and a velocity estimation precision of 1.40 km/h (std). These precisions have been measured by comparing the LiDAR position with a ground truth reference provided by the RTK GNSS receivers during the same data acquisition step, keeping only the intervals where both sensors had RTK Fixed integer-level precision. Also, it must be noted that the manual temporal annotation of the VUT crossing the lane marking, representing the TSM instant, has a 1-second resolution.

These uncertainties can be represented as a rectangle around the TLC marker in **Figure 4** graphs, and the statistical significance of the experiment lies in the margin between the measurements and the R79 distance thresholds. In this regard, it is interesting to note that, most of the identified experiments with potential Critical Distance Overshooting where the lane change was not suppressed took place in a low delta-v area, where the R79 curve is flat, and its precision is less relevant.

Given these precision figures, we can claim 3-sigma or 99% confidence interval about the Critical Distance Overshooting for experiments 16, 16b and 16d, but we cannot have the same certainty for experiments 10, 10b and 10d, even though they have 2-sigma significance.

These experiments prove the validity of the LiDAR-based measurements in detecting R79 Critical Distance Overshoots on public roads, defining their fitness for type approval.

shows the final categorization of the experiments taking into account the measurement uncertainty determined for the LiDAR based system, highlighting the 3 detected Critical Distance Overshoots with 99% Confidence Interval as a subset of the Critical Completed experiments.

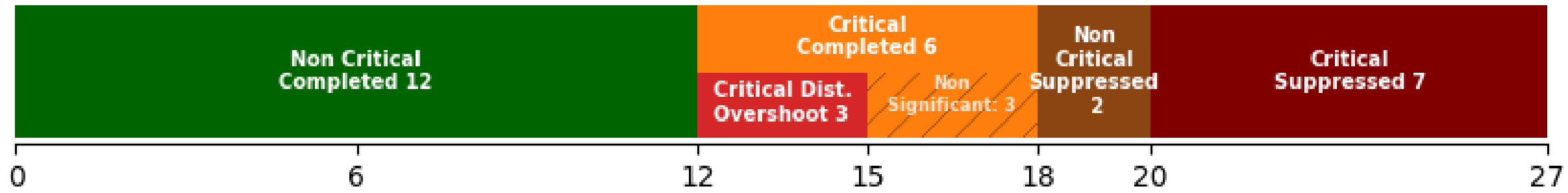

**Figure 5: Count by final category of all the experiments divided in the four conditions, highlighting the detected Critical Distance Overshoots with 99% Confidence Interval.**

**Figure 4** shows how all the Critical Completed experiments, where the ALC was not supressed despite the R79 Critical Distance Overshoot, were characterized by small speed differences (low values of delta-v). Equation (1) is defined assuming the TO vehicle being faster than the VUT, and enforces the rationale imposing a time headway of 1s for the TO vehicle, limiting its required deceleration within safe boundaries. Whenever deceleration is not necessarily needed because of the low speed difference, as in the 6 detected overshoots, the 1s headway limit should still be respected, as it is depicted by the R79 boundary curve in **Figure 4.**

This edge case may result in a failed detection of the critical threshold by the system. Nevertheless, none of the Critical Completed ALC manoeuvres were perceived as dangerous by the testing personnel. These Critical Distance Overshoots are all found in the areas of the operating domain outside of what prescribed to the Type Approval tests under UNECE R79, which only mandate a single Critical manoeuvre abort test on a proving ground with VUT speed under 100 km/h. However, on public roads many other combinations can occur, and the system should be robust to domain variations.

## CONCLUSIONS

This paper presents a methodology for the independent testing of commercial Assisted Lane Change systems, focusing on systems type approved according to UNECE Regulation No. 79. The proposed approach uses LiDAR-based measurements as an alternative to conventional RTK-GNSS-based instrumentation, enabling coordinated public road testing of geo-fenced systems without manufacturers' support and limitation by high precision GNSS data (e.g. RTK).

The test methodology was applied to 27 Assisted Lane Change experiments carried out by type-approved passenger cars during a dedicated on-road campaign encompassing line changes performed at different relative distances and speeds by following a full factorial design. Comparison with the R79 Critical Distance threshold identified six completed with an R79 Critical Distance Overshot, where the ALC was completed even if the approaching vehicle was closer than the prescribed value. After considering spatial and temporal measurement uncertainty, 3 out of 6 experiments have a 99% Confidence Interval measurement of the Critical Distance Overshot.

These results show that public-road testing with LiDAR based system is precise enough to detect critical system behaviour under real word operating regions that may be insufficiently represented during conventional type-approval testing. The proposed methodology could therefore support independent market surveillance and in-service monitoring assessment of commercially available ALC systems under real-world conditions.

Despite the limitations associated with testing a single vehicle and with the measurement and temporal annotation precision, this work demonstrated the feasibility of independently assessing commercial ALC systems under real-world public-road conditions. Future work will focus on improving the accuracy of the LiDAR system and temporal annotations, moving towards multi-sensor fusion and automated event detection. This would develop the proposed approach into a repeatable procedure for market surveillance, in-service monitoring and the evaluation of future revisions of R79.

**ACKNOWLEDGMENTS**
The authors acknowledge the technical and operational support of the Netherlands Vehicle Authority (RDW) and IVEX NV (Heverlee, Belgium).
The authors acknowledge the help of Giulia Morandin, Jessica Leoni, Sandor Vass and Giovanni Grieco for their help in proof-reading and improving the quality of the manuscript.
Authors used AI tools to help with this work: ChatGPT was used for help in the Literature Research and general text proof-reading and quality improvement, while Codex and Claude helped generate the data analysis software. The authors thoroughly assessed the AI output, and assume full responsibility for the accuracy, interpretation, and results of all this work.

**AUTHOR CONTRIBUTIONS**
The authors confirm contribution to the paper as follows: study conception and design: Akos Kriston, Marcello Cellina, Biagio Ciuffo; experimental setup: Marcello Cellina, Antonio Migneco, Stefano Favelli; data collection: Marcello Cellina, Antonio Migneco, Davide Maggi, Stefano Favelli, Fabrizio Re, Andrea Nuovo, Fabrizio Minarini; analysis and interpretation of results: Akos Kriston, Marcello Cellina, Riccardo Dona; draft manuscript preparation: Marcello Cellina. All authors reviewed the results and approved the final version of the manuscript.

**DECLARATION OF CONFLICTING INTERESTS**
The authors declared no potential conflicts of interest with respect to the research, authorship, and/or publication of this article.

**FUNDING**
The authors disclosed no financial support for the research, authorship, and/or publication of this article.